\documentclass[3p,twocolumn]{elsarticle}

\usepackage{lineno,hyperref}
\usepackage{booktabs}
\usepackage{ragged2e}
\usepackage{xcolor}
\usepackage{nicefrac}
\usepackage{placeins}

\usepackage{algorithm}
\usepackage{mathtools}
\usepackage{amssymb}
\usepackage{url}
\usepackage{amsmath}
\usepackage{graphicx}
\usepackage{subcaption}
\usepackage{array}
\usepackage{algorithm}
\usepackage[noend]{algpseudocode}
\usepackage[misc]{ifsym}
\usepackage{fixltx2e}
\usepackage{tikz}
\usepackage{color}
\usetikzlibrary{positioning}
\usepackage{float}
\usepackage{chngcntr}

\modulolinenumbers[5]
\newcolumntype{P}[1]{>{\centering\arraybackslash}p{#1}}

\journal{NeuroImage}

\begin{document}

\begin{frontmatter}

\title{Analyzing the effect of \textit{APOE} on Alzheimer's disease progression using an event-based model for stratified populations}
\author[add1]{Vikram Venkatraghavan\corref{mycorrespondingauthor}}
\cortext[mycorrespondingauthor]{Corresponding author}
\ead{v.venkatraghavan@erasmusmc.nl}
\author[add1]{Stefan Klein}
\author[add3]{Lana Fani}
\author[add1]{Leontine S. Ham}
\author[add1]{Henri Vrooman}
\author[add3,add4]{\mbox{M. Kamran Ikram}}
\author[add1,add2]{Wiro J. Niessen}
\author[add1]{Esther E. Bron}
\author{\mbox{for the Alzheimer's Disease Neuroimaging Initiative} \fnref{fn1}}
\fntext[fn1]{Data used in preparation of this article were obtained from the Alzheimer's Disease Neuroimaging Initiative (ADNI) database (adni.loni.usc.edu). As such, the investigators within the ADNI contributed to the design and implementation of ADNI and/or provided data but did not participate in analysis or writing of this report. A complete listing of ADNI investigators can be found at: \url{http://adni.loni.usc.edu/wp-content/uploads/how_to_apply/ADNI_Acknowledgement_List.pdf}}

\address[add1]{Biomedical Imaging Group Rotterdam, Department of Radiology \& Nuclear Medicine, Erasmus MC, University Medical Center Rotterdam, The Netherlands}
\address[add2]{Quantitative Imaging Group, Dept. of Imaging Physics, Faculty of Applied Sciences, Delft University of Technology, Delft, The Netherlands}
\address[add3]{Department of Epidemiology, Erasmus MC, University Medical Center Rotterdam, The Netherlands}
\address[add4]{Department of Neurology, Erasmus MC, University Medical Center Rotterdam, The Netherlands}
\begin{abstract}

\noindent Alzheimer's disease (AD) is the most common form of dementia and is phenotypically heterogeneous. \textit{APOE} is a triallelic gene which correlates with phenotypic heterogeneity in AD. In this work, we determined the effect of \textit{APOE} alleles on the disease progression timeline of AD using a discriminative event-based model (DEBM). Since DEBM is a data-driven model, stratification into smaller disease subgroups would lead to more inaccurate models as compared to fitting the model on the entire dataset. Hence our secondary aim is to propose and evaluate novel approaches in which we split the different steps of DEBM into group-aspecific and group-specific parts, where the entire dataset is used to train the group-aspecific parts and only the data from a specific group is used to train the group-specific parts of the DEBM. We performed simulation experiments to benchmark the accuracy of the proposed approaches and to select the optimal approach. Subsequently, the chosen approach was applied to the baseline data of $417$ cognitively normal, $235$ mild cognitively impaired who convert to AD within $3$ years, and $342$ AD patients from the Alzheimer’s Disease Neuroimaging Initiative (ADNI) dataset to gain new insights into the effect of \textit{APOE} carriership on the disease progression timeline of AD. In the risk group (with \textit{APOE} $\varepsilon 3-\varepsilon4$ or $\varepsilon 4-\varepsilon4$), the model predicted with high confidence that CSF Amyloid $\beta$ and the cognitive score of Alzheimer's Disease Assessment Scale (ADAS) are early biomarkers. Hippocampus was the earliest volumetric biomarker to become abnormal, with the molecular layer of the hippocampus becoming abnormal even before the CSF Phosphorylated Tau (PTAU) biomarker. In the neutral group (with \textit{APOE} $\varepsilon 3-\varepsilon 3$) the model predicted a similar ordering among CSF biomarkers. However, the volume of the fusiform gyrus was identified as one of the earliest volumetric biomarker along with the hippocampus subparts of Fissure and Parasubiculum. While the findings in the risk and neutral groups fit the amyloid cascade hypothesis, the finding in the protective group (with \textit{APOE} $\varepsilon 2-\varepsilon2$ or $\varepsilon 2-\varepsilon3$) did not. The model predicted, with relatively low confidence, CSF Neurogranin as one of the earliest biomarkers along with cognitive score of Mini-Mental State Examination (MMSE). Amyloid $\beta$ was found to become abnormal after PTAU. The presented models could aid understanding of the disease, and in selecting homogeneous group of presymptomatic subjects at-risk of developing symptoms for clinical trials. 

\end{abstract}


\end{frontmatter}

\section{Introduction} \label{sec:intro}

Dementia affects roughly $5\%$ of the world's elderly population of whom $60-70\%$ are affected by Alzheimer's Disease (AD), which is the most common form of dementia~\citep{WHO:2017}. There are several neurobiological subtypes of AD~\citep{Ferreira:2020} and each subtype potentially needs a different strategy to prevent or slow the progression of AD. Understanding the pathophysiological processes in AD is thus crucial for selecting novel preventive or therapeutic targets for clinical trials of disease modifying treatments, identifying target groups for such trials and tracking the disease progression in patients. 

While several studies have looked into the pathophysiology of AD~\citep{Jack:2013, Bloom:2014, Weigand:2019}, it is still not completely understood. Although it has been observed that AD is phenotypically heterogeneous~\citep{Murray:2011, Au:2015, Patterson:2018} with potentially different pathways for disease progression, these pathways remain unclear. There is hence a need to understand the phenotypic heterogeneity in AD while leveraging neuroimaging, fluid and cognitive biomarkers. 

\textit{APOE} is a triallelic gene in which the $\varepsilon2$ allele reduces the risk of AD~\citep{vanderlee:2018}, the $\varepsilon3$ allele acts as a reference allele and the $\varepsilon4$ allele is a major genetic risk factor of AD~\citep{Saunders:1993,Kim:2009, Emmanuelle:2011}. \textit{APOE} has been shown to correlate with phenotypic heterogeneity in AD~\citep{Sandra:2019}. Hence we hypothesize that the pathophysiology of AD can be better understood when considering the effect of \textit{APOE} carriership on biomarker changes.

In the context of data-driven methods for understanding AD pathophysiology, disease progression models have been used to study the trajectories of individual biomarkers~\citep{Jedynak:2012, Schiratti:2015, Lorenzi:2017} as well as their progression with respect to each other~\citep{Fonteijn:2012,Vikram:2017,Young:2014,Huang:2012}. Unlike typical machine learning approaches, these models are interpretable by design and provide insight for understanding the mechanisms of disease progression. Event-based models (EBMs) are a class of such interpretable disease progression models that estimate the timeline of neuropathologic change during AD progression using cross-sectional data~\citep{Fonteijn:2012,Vikram:2019}.

Our primary aim is to use the discriminative event-based model (DEBM), which was shown to be more accurate than previously proposed EBMs~\citep{Vikram:2019}, to understand the effect of different \textit{APOE} alleles on the disease timeline of AD. To shed light on different aspects of neurodegeneration and identify the earliest brain regions affected, we included commonly studied cerebrospinal fluid (CSF) biomarkers, cognitive scores, and volumetric biomarkers from neuroimaging including those of hippocampus subfields. Hippocampus  subfields were included in our model because while Hippocampus neuronal dysfunction is quite commonly studied to be affected in AD, it is not a homogenous structure, and subfields could be affected differentially due to \textit{APOE} carriership~\citep{Mueller:2009}.

The default approach for estimating the disease progression timeline would be to stratify the population based on their \textit{APOE} $\varepsilon 2-4$ carrier status and independently train the DEBM model on the stratified populations~\citep{Young:2014}. However, since DEBM is a data-driven model, stratification into smaller groups would lead to less accurate models than those obtained by the original method on the entire dataset. Hence our secondary aim is to propose and evaluate a novel approach in which we split the different steps of DEBM into group-aspecific and group-specific parts, where the entire dataset is used to train the group-aspecific parts and only the data from a specific group is used to train the group-specific parts of the DEBM. We present two different variations of this approach and we hypothesize that the optimal split of the DEBM steps into the group-aspecific and group-specific parts would result in better accuracy of the estimated disease progression timeline. Since the ground-truth timelines are unknown in a clinical setting, we evaluate the accuracy of the proposed variations using simulation experiments and we select the optimal method for the analysis on the effect of \textit{APOE} on the AD progression timeline on patient data.

To summarize, our contributions in this paper include proposing and evaluating a novel approach for using DEBM in stratified populations and estimating a comprehensive timeline of AD progression, in terms of biomarker changes, in the presence of different \textit{APOE} alleles.


\section{Methods} \label{sec:methods}

\begin{figure*}[t!]
\centering
\includegraphics[width=1\textwidth]{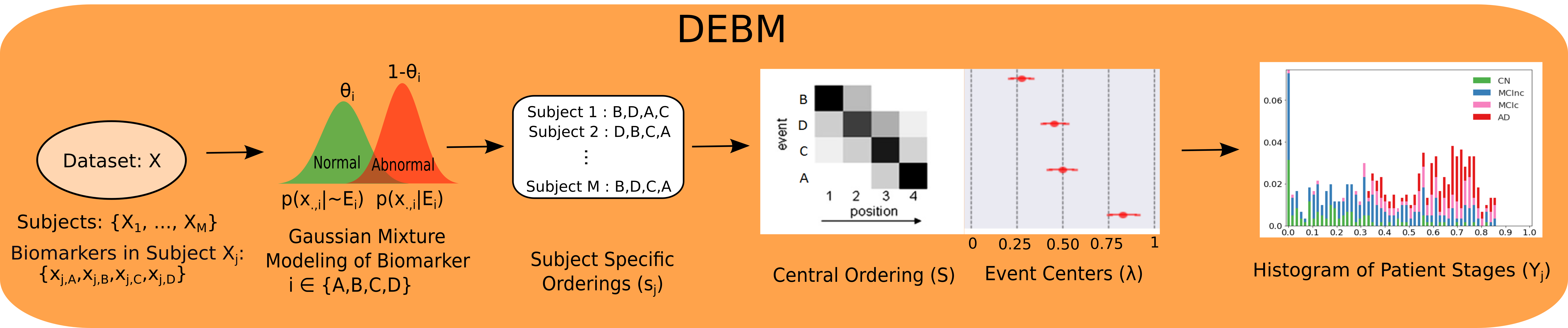}
\caption{Overview of the steps involved in DEBM. Input for the DEBM model is a cross-sectional dataset $X$ with $M$ subjects and various biomarkers ($A$,$B$,$C$ and $D$) representing different aspects of neuro-degeneration. Using Gaussian mixture modeling (GMM), mixing parameters $(\theta_i)$ and probability density functions of normal $(p(x_{\cdot,i}|\neg E_i))$ and abnormal $(p(x_{\cdot,i}|E_i))$ levels are estimated for each biomarker. This is followed by the estimation of subject-specific orderings $(s_j)$, for each subject in the dataset. Disease progression timeline consisting of central ordering $(S)$ and event-centers $(\lambda)$ are estimated based on these subject-specific orderings. Based on the constructed disease progression timeline, patient stages $(\Upsilon_j)$ of subjects in an independent test-set can be estimated.}
\label{fig:DEBM_Illustration}
\end{figure*}

An introduction to the DEBM model~\citep{Vikram:2019} is provided in Section~\ref{ssec:debm}. In Section~\ref{ssec:coupled} we propose our novel approach for using DEBM in stratified populations with its two variations.

\subsection{Discriminative event-based modeling}  \label{ssec:debm}

In a cross-sectional dataset $(X)$ of $M$ subjects, including cognitively normal individuals (CN), subjects with mild cognitive impairment (MCI) and patients with AD, let $X_j$ denote a measurement of biomarkers for subject $j\in\left[1,M\right]$, consisting of scalar biomarker values $x_{j,i}$ for $i \in \left[1,N \right]$. $x_{\cdot,i}$ denotes the $i$-th biomarker for any unspecified $j$. DEBM estimates the posterior probabilities of individual biomarkers being abnormal. These posterior probabilities are used to estimate the ordering of biomarker changes for each subject independently. The central ordering and disease progression timeline for the entire dataset are estimated based on these subject-specific orderings. The resulting disease progression timeline is used for assessing the severity of disease in an individual based on his/her biomarker values. Figure~\ref{fig:DEBM_Illustration} shows the different steps involved in DEBM.

\textbf{Step 1 - Mixture Modeling:} As AD is characterized by a cascade of neuropathological changes that occurs over several years, presymptomatic CN subjects can have some abnormal biomarker values. On the other hand, in atypical AD subjects, a proportion of biomarkers may still have normal values, especially in patients at an early disease stage. Hence clinical labels cannot directly be propagated to individual biomarkers to label normal and abnormal biomarker values. We shall refer to this as biomarker label noise in the rest of the paper. In order to estimate the posterior probabilities of individual biomarkers being abnormal, DEBM, similar to previously proposed EBMs~\cite{Fonteijn:2012,Huang:2012,Young:2014}, fits a Gaussian mixture model (GMM) to construct the normal / pre-event probability density function (PDF), $p(x_{\cdot,i} | \neg E_i)$, and abnormal / post-event PDF, $p(x_{\cdot,i} | E_i)$. Event $E_i$ in this notation is used to denote the corresponding biomarker becoming abnormal and $\neg E_i$ denotes the corresponnding biomarer being normal. The aforementioned PDFs can be expressed as:

\begin{equation}
p(x_{\cdot,i} | \neg E_i) = \mathcal{N}(\mu_{i,\neg E};\sigma_{i,\neg E})
\end{equation}

\begin{equation}
p(x_{\cdot,i} | E_i) = \mathcal{N}(\mu_{i, E};\sigma_{i, E})
\end{equation}

\noindent Where, $\mathcal{N(\mu,\sigma)}$ is the normal distribution with mean $\mu$ and standard deviation $\sigma$. 

\begin{figure}[t!]
\centering
\includegraphics[width=0.5\textwidth]{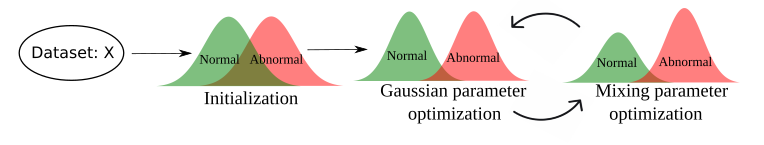}
\caption{Overview of GMM optimization in DEBM.}
\label{fig:GMM_Illustration}
\end{figure}

For estimating these parameters robustly in the presence of biomarker label noise, the normal and abnormal PDF estimates are first initialized using the mean and standard deviations after truncating the overlapping tails of the observed distributions in CN and AD subjects. This can be observed in Figure~\ref{fig:GMM_Illustration}, where the initialization is performed only based on the non-overlapping parts of green and red curves, while the overlapping part is left out to account for biomarker label noise. At this stage of GMM initialization, MCI subjects are left out as well, because it is unsure a priori whether their biomarkers are normal or abnormal. The resulting initialized PDFs are denoted as $\widehat{p}(x_{\cdot,i}|\neg E_i)$) and $\widehat{p}(x_{\cdot,i}|E_i)$. 

This is followed by an alternating GMM maximum likelihood optimization scheme until both the Gaussian parameters as well as the mixing parameters converge. All the subjects, including MCI, are used for GMM optimization. After convergence, these Gaussians are used to represent the PDFs $p(x_{\cdot,i}|\neg E_i)$ and $p(x_{\cdot,i}|E_i)$. The mixing parameters $(\theta_i)$ are used as prior probabilities to convert these PDFs to posterior probabilities $p(\neg E_i | x_{\cdot,i})$ and $p(E_i | x_{\cdot,i})$. Figure~\ref{fig:GMM_Illustration} shows an overview of this optimization scheme.

\noindent \textbf{Step 2 - Subject-specific Orderings:} $p(E_i | x_{j,i}) \forall i$ are used to estimate the subject-specific orderings $s_j$. $s_j$ is established such that:

\begin{multline}
\label{eq:s_j}
s_j \ni p(E_{s_j(1)} | x_{j,s_j(1)}) > ... \\ > p(E_{s_j(N)} | x_{j,s_j(N)})
\end{multline}

\textbf{Step 3 - Central Ordering:} DEBM computes the central event ordering $S$ from the subject-specific estimates $s_j$. To describe the distribution of $s_j$, a generalized Mallows model is used~\citep{Fligner:1988}. The central ordering is defined as the ordering that minimizes the sum of distances to all subject-specific orderings $s_j$, with probabilistic Kendall's Tau being the distance measure~\citep{Vikram:2019}. While $S$ denotes the sequence of biomarker events, the relative position of these events (event-centers) in a normalized scale of $\left[ 0,1\right]$ is denoted by the vector $\lambda$. The pair $\{S,\lambda \}$ together forms a disease progression timeline.

\textbf{Step 4 - Patient Staging:} Once the disease progression timeline is created, subjects in an independent test set $(T)$ can be placed on this timeline to estimate disease severity. This is achieved by converting the biomarker values of the test subjects to posterior probabilities $p(E_i | x_{j,i})$, $\forall j \in T$. These can be used to estimate disease severities in test subjects by first estimating the conditional distribution $p(i|S,X_j)$, which estimates the probability that the first $i$ events of $S$ have occurred for a test-subject and the rest are yet to occur.

\begin{multline}
\label{eq:pstage2}
p\left(i | S , X_j \right) \propto \prod_{l=1}^i p \left( E_{S(l)} | x_{j,S(l)} \right) \times \\ \prod_{l=i+1}^N p \left( \neg E_{S(l)} | x_{j,S(l)} \right)
\end{multline}

The patient stage of a test subject ($\Upsilon_j$) is defined as the expectation of $\lambda (i)$ with respect to the conditional distribution $p(i|S,X_j)$.

\begin{equation}
\label{eq:pstage}
\Upsilon_j = \frac{ \sum_{i=1}^N \lambda(i)  p(i | S,X_j)}{\sum_{i=1}^N p(i | S,X_j)}
\end{equation}

\subsection{Group-specific and group-aspecific parts of DEBM} \label{ssec:coupled} 

We propose extensions of DEBM for stratified populations, i.e., when the dataset $X$ can be subdivided in groups $g \in \left[1,G \right]$, based on, e.g., genotype or phenotype of the subjects. Since DEBM is a data-driven model, data stratification into smaller groups would lead to more inaccurate models~\citep{Vikram:2019}. To obtain better DEBM accuracies in such scenario, we propose to co-train DEBM for estimating disease timelines $\forall g$ by splitting DEBM into group-aspecific and group-specific parts. The group-aspecific parts of DEBM are estimated using the entire dataset and group-specific parts are estimated for each group independently. 

We first discuss the default way of independently training DEBM in the different groups and then propose two different approaches for splitting DEBM into group-aspecific and group-specific parts.

\textbf{Approach 1: Independent DEBM} 

\begin{figure}[t!]

\begin{subfigure}[b]{0.5\textwidth}
    \centering
\begin{tikzpicture}
  \node (img)  {\includegraphics[width=0.9\textwidth]{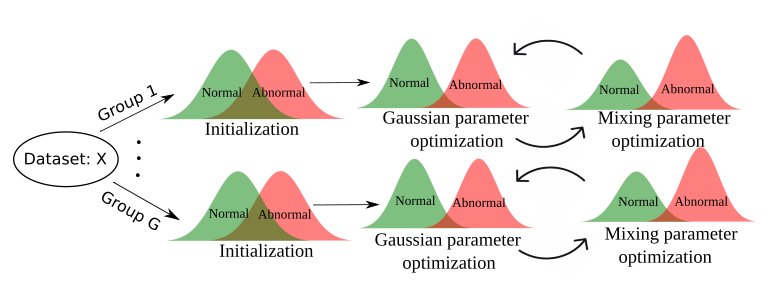}};
\end{tikzpicture}
  \caption{GMM in independent DEBM}
  \end{subfigure}

\begin{subfigure}[b]{0.5\textwidth}
    \centering
\begin{tikzpicture}
  \node (img)  {\includegraphics[width=0.9\textwidth]{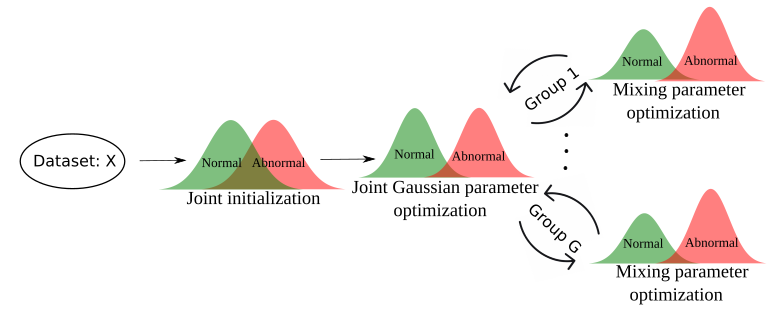}};
\end{tikzpicture}
  \caption{GMM in Coupled DEBM}
\end{subfigure}

    \begin{subfigure}[b]{0.5\textwidth}
    \centering
\begin{tikzpicture}
  \node (img)  {\includegraphics[width=0.9\textwidth]{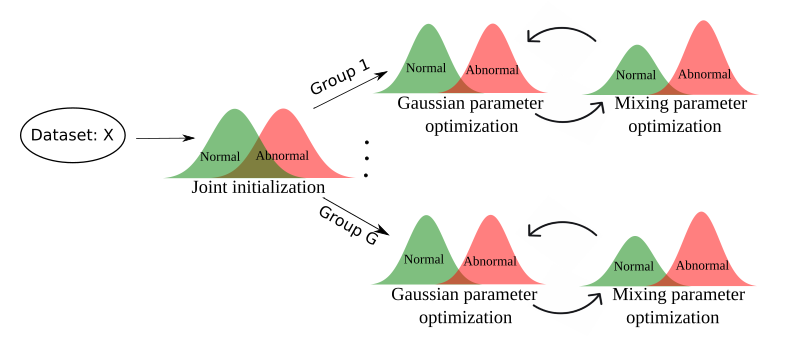}};
\end{tikzpicture}
  \caption{GMM in Co-init DEBM}
  \end{subfigure}

\caption{Overview of GMM optimization strategies in the different approaches for DEBM analysis in stratified populations. (a) The default approach in which GMM in each group is trained independently. (b) GMM in coupled DEBM, where the different groups share the Gaussian parameters, but the mixing parameters are estimated independently. (c) GMM in co-init DEBM in which the different groups are jointly initialized before the GMM optimization, but the optimization is done independently for each group.}
\label{fig:GMMIllustration2}
\end{figure}

In this default approach, each group is considered as an independent dataset and the disease progression timeline in each group is estimated independently. GMM in such a scenario is illustrated in Figure~\ref{fig:GMMIllustration2}a.

\textbf{Approach 2: Coupled DEBM} 

\begin{equation}
\label{eq:split1}
 \text{DEBM} \rightarrow
    \begin{cases}
      p(x_{\cdot,i}|\neg E_i),p(x_{\cdot,i}|E_i) & \text{group-aspecific} \\
      \theta_{i,g},\{S_g$, $\lambda_g\}, & \text{group-specific} \\
    \end{cases}   
\end{equation}

\noindent In this approach, we assume that the different groups share the normal and abnormal PDFs, but the ordering in which these biomarkers become abnormal are different. 
The mixing parameters ($\theta_{i,g}$) are considered as group-specific part of the DEBM algorithm because the proportion of subjects with normal and abnormal biomarker values in each group $g$ is correlated with the position of the biomarker along the ordering $S_g$, which we expect to be different in each group.

Hence, in our approach, we modify the alternating GMM optimization scheme to jointly optimize the GMM parameters of multiple groups. First, the GMM algorithm is initialized without considering the groups, as explained in Section~\ref{ssec:debm}. Secondly, as with the default DEBM, Gaussian parameters and mixing parameters are alternately optimized. In contrast in coupled DEBM, the Gaussian parameters are estimated jointly for all groups, while mixing parameters are estimated separately for each group. This has been illustrated in Figure~\ref{fig:GMMIllustration2}b. 

Once the GMM optimization has been performed, $S_g$ and $\lambda_g$ are estimated in each group. Patient staging $(\Upsilon_j)$ of the test-subjects in group $g$ are computed based on the disease progression timeline $\{ S_g, \lambda_g \}$.

\textbf{Approach 3: Co-init DEBM} 

\begin{equation}
\label{eq:split2}
 \text{DEBM} \rightarrow
    \begin{cases}
      \widehat{p}(x_{\cdot,i}|\neg E_i),\widehat{p}(x_{\cdot,i}|E_i) & \text{group-aspecific}\\
       p_g(x_{\cdot,i}|\neg E_i),p_g(x_{\cdot,i}|E_i) & \text{group-specific}\\
      \theta_{i,g},\{S_g, \lambda_g\} & \text{group-specific} \\
    \end{cases}   
\end{equation}

In this approach, we assume that the different groups do not share the normal and abnormal PDFs, but that they are close to each other. Hence, in co-init DEBM, we relax the constraint on $p(x_{\cdot,i}|\neg E_i)$ and $p(x_{\cdot,i}|E_i)$ and instead consider the initialized values of normal and abnormal PDFs ($\widehat{p}(x_{\cdot,i}|\neg E_i)$ and $\widehat{p}(x_{\cdot,i}|E_i)$) to be group-aspecific part of DEBM. We estimate $p_g(x_{\cdot,i}|\neg E_i)$ and $p_g(x_{\cdot,i}|E_i)$ independently for each group. This is illustrated in Figure~\ref{fig:GMMIllustration2}c.

As with the previous approach, $S_g$, $\lambda_g$ and the patient staging of the test-subjects in group $g$ are computed independently for each group.

\section{Experiments}  \label{sec:exp}


Section~\ref{ssec:simexp} describes the experiments to evaluate the proposed DEBM approaches on a stratified population. Since ground-truth orderings are unknown in real clinical data, we use simulated datasets for evaluating the methods. After evaluating the proposed approaches, we select the best approach for analyzing the effect of \textit{APOE} on AD progression using subjects from the Alzheimer's Disease Neuroimaging Initiative (ADNI) database. Section~\ref{ssec:adniexp} descibes the details of these experiments.

\subsection{Simulation Experiments}  \label{ssec:simexp}

We used the framework developed by~\cite{Young:2015b} for simulating cross-sectional data consisting of scalar biomarker values for CN, MCI and AD subjects in two groups. In this framework, disease progression in a subject is modeled by a series of biomarker changes representing the temporal cascade of biomarker abnormality as estimated by an EBM. Individual biomarker trajectories are represented by sigmoids varying from the biomarker's normal value to its abnormal value. To account for inter-subject variability, the normal and abnormal values for different subjects are drawn randomly from Gaussian distributions.

The simulation dataset used in our experiments are based on a set of seven biomarkers as described in the simulation experiments of~\cite{Vikram:2019}. The simulated datasets were stratified into two groups, with each group having its own distinct disease progression patterns. There are two ways in which the progression of disease in the groups can differ: 1. difference in ground-truth orderings $S_1$ and $S_2$; 2. difference in the abnormal biomarker PDFs in the two groups \textit{i.e.} $p_1(x_{\cdot,i}|E_i)$ and $p_2(x_{\cdot,i}|E_i)$. Each of these differences could affect the accuracy of the proposed approaches. Hence, we evaluated the proposed approaches in the presence of each of these differences. Normalized Kendall's Tau distance between the estimated ordering ($S$) and the ground-truth ordering ($S_{gt}$) was used as an evaluation measure in these experiments:

\begin{equation}
\label{eq:E1}
\varepsilon_{S} = \nicefrac{K(S,S_{gt})}{\binom{N}{2}}
\end{equation}

\noindent where $K(A,B)$ is the number of swaps required to obtain ordering B from ordering A. 

The normalization ensures that $\varepsilon_{S}$ falls in the range $\left[0,1\right]$, with $0$ as the distance when  the two orderings are the same, and $1$ as the distance when the two orderings are the reverse of each other.

\textbf{Experiment 1:} The first simulation experiment studied the effect of the difference in ordering between the two groups. The ordering in the first group (Group $1$) was fixed and the ordering in the second group (Group $2$) was selected randomly such that the normalized Kendall's Tau distance between the two groups was a fixed number, say $\varepsilon_O$. $\varepsilon_O$ was varied from $0$ to $1$ in steps of $0.2$. The number of subjects in Group $2$ was kept constant at $900$. The number of subjects in Group $1$ was varied from $100$ to $900$ in steps of $200$, to study how the different approaches perform in small as well as large groups. The normal and abnormal biomarkers levels in the two groups were sampled from the same Gaussian distribution for this experiment. We generated $50$ random repetitions of the simulated datasets, and reported mean and standard deviation of $\varepsilon_{S}$ for independent DEBM, coupled DEBM, and co-init DEBM in groups $1$ and $2$.

\textbf{Experiment 2:} This experiment studied the performance of the proposed approaches with the $\mu_{g,i,E}$ parameter of the $p_g(x_{\cdot,i}|E_i)$ distribution being different in the two groups. $\mu_{1,i,E}$ was fixed, and $\mu_{2,i,E}$ was varied such that the difference $\mu_{2,i,E} - \mu_{1,i,E}$ ($\varepsilon_G$) was one of $\{-0.2d, 0, +0.2d\}$ where $d = \mu_{1,i,E} - \mu_{1,i,\neg E}$. $0$ is considered the reference level, where the abnormal Gaussians are the same in the two groups. $\mu_{g,i,\neg E}$ were kept the same in the two groups. Hence, when $\varepsilon_G=-0.2d$, the abnormal biomarker levels are closer to the normal biomarker levels in Group 2 than in Group 1. This results in Group 2 biomarkers being weaker than their Group 1 counterparts when $\varepsilon_G=-0.2d$ and stronger when $\varepsilon_G=+0.2d$. The number of subjects in Group $2$ was kept a constant at $900$, while the subjects in Group $1$ increased from $100$ to $900$. $\varepsilon_O$ between the two groups was fixed at $0.4$. We again generated $50$ random repetitions of the simulated datasets, and reported mean and standard deviation of $\varepsilon_{S}$ for coupled DEBM, co-init DEBM and DEBM.

These experiments were used to evaluate the different approaches mentioned in Section~\ref{sec:methods} and select the best method for analyzing the effect of \textit{APOE} alleles in AD progression.

\subsection{Studying the effect of \textit{APOE}}  \label{ssec:adniexp}

We considered the baseline measurements from $417$ CN, $235$ MCI converters and $342$ AD subjects in ADNI1, ADNIGO and ADNI2 studies\footnote{The ADNI was launched in 2003 as a public-private partnership, led by Principal Investigator Michael W. Weiner, MD. The primary goal of ADNI has been to test whether serial magnetic resonance imaging (MRI), positron emission tomography (PET), other biological markers, and clinical and neuropsychological assessment can be combined to measure the progression of mild cognitive impairment (MCI) and early Alzheimers disease (AD). For up-to-date information, see www.adni-info.org.}. The MCI converters are subjects who had MCI at baseline but converted to AD within 3 years of baseline measurement. We excluded subjects with significant memory concerns (without a diagnosis of AD or MCI) and MCI non-converters in our experiments to select a more phenotypically homogeneous group of subjects with prevalent or incident AD. In each of the experiments, the dataset was divided into three groups (protective, neutral and risk) based on the subject's \textit{APOE} carriership.  The protective group consisted of subjects with \textit{APOE} $\varepsilon2,2$ and $\varepsilon2,3$~\citep{vanderlee:2018}. The neutral group consisted of subjects with \textit{APOE} $\varepsilon3,3$. The risk group consisted of subjects with \textit{APOE} $\varepsilon3,4$ and $\varepsilon4,4$~\citep{Saunders:1993,Kim:2009, Emmanuelle:2011}. Subjects with \textit{APOE} $\varepsilon2,4$ (n=34) were not included in either group because of the presence of both protective and risk alleles.

Subject demographics and their \textit{APOE} carrierships are summarized in Table~\ref{tab:Demographics}. The modalities considered were structural imaging (T1-weighted MRI) biomarkers, biomarkers extracted from cerebrospinal fluid (CSF), and cognitive biomarkers. Details of the MRI acquisition protocols of ADNI can be found in~\cite{Jack:2008, Jack:2015}.

\begin {table}
\footnotesize
\begin{center}
 \begin{tabular}{|P{2cm}| P{1.3cm} P{1.3cm} P{1.4cm}|} 
 \hline
 \multicolumn{4}{|c|}{Demographics} \\ \hline
 Diagnosis & CN  & MCIc & AD \\ \hline
 $n$ &  $417$ & $235$ & $342$ \\ \hline
 \textit{APOE} $2\star$/$33$/$\star 4$ & $57$/$244$/$110$  & $6$/$66$/$156$ & $12$/$101$/$219$ \\ \hline
 Sex M/F &  $209/208$ & $145/90$ & $189/153$ \\ \hline
 Age [yrs.] ($\mu\pm\sigma$) &  $74.8\pm5.7$ & $73.7 \pm 7.0$  & $75.0\pm7.8$ \\ \hline
   Edu [yrs.] ($\mu\pm\sigma$) & $16.3\pm2.7$ & $15.9 \pm 2.7$ & $15.2\pm3.0$ \\ \hline
\end{tabular}
\end{center}
\caption{Demographics for the used population. $2\star$ represents the subjects with protective \textit{APOE} alleles $\varepsilon2,2$ and $\varepsilon2,3$. $33$ represents the  subjects with reference \textit{APOE} allele $\varepsilon3,3$. $\star4$ represents the subjects with risk \textit{APOE} alleles $\varepsilon3,4$ and $\varepsilon4,4$. Subjects with both protective and risk allele $\varepsilon2,4$ were excluded from this study (n=$34$). Edu. is an abbreviation used for Education.}
\label{tab:Demographics} 
\end{table}

Imaging biomarkers were estimated from T1-weighted MRI scans analysed with FreeSurfer software v6.0 cross-sectional stream and outputs were visually checked. We assumed a symmetric pattern of atrophy in AD and averaged imaging biomarkers between the left and right hemisphere.

\noindent \textbf{Experiment 3:} For this experiment, the selected imaging biomarkers were: volumetric measures of the hippocampus, entorhinal cortex, fusiform gyrus, middle-temporal gyrus and precuneus, together with whole brain volume and ventricles~\citep{Archetti:2019, Frisoni:2010, Vemuri:2010}. The selected CSF based biomarkers were: CSF concentrations of Amyloid-β (ABETA), total Tau (TAU) and phosphorylated Tau (PTAU) proteins~\citep{Blennow:2003,Blennow:2010}, Neurogranin~\citep{Thorsell:2010} and Neurofilament light chain~\citep{Mei:2019,deWolf:2020}. Mini mental state examination (MMSE) and Alzheimer's Disease Assessment Scale - Cognitive (13 items) (ADAS13) were used as cognitive biomarkers. The availability of these multimodal biomarkers in the ADNI database is summarized in Table 2.

\begin {table}
\footnotesize
\begin{center}
 \begin{tabular}{|P{1.6cm}| P{1.2cm} P{1.4cm} P{1.4cm}|} 
 \hline
 \multicolumn{4}{|c|}{Biomarker Availability} \\ \hline
 Biomarker & Protective $(N=75)$ & Neutral $(N=411)$ & Risk $(N=485)$ \\ \hline
 Imaging & $74$ & $408$ & $481$ \\ 
 ABETA & $57$ & $301$ & $357$ \\ 
 PTAU & $57$  & $301$ & $357$ \\
 TAU & $57$  & $299$ & $348$ \\ 
 NG & $21$  & $113$ & $131$ \\ 
 NFL & $23$ & $118$ & $137$ \\ 
 MMSE & $75$ & $411$ & $485$ \\
 ADAS & $74$ & $410$ & $477$ \\ \hline
\end{tabular}
\end{center}
\caption{Biomarker availability in number of subjects in the protective, neutral and risk groups.}
\label{tab:BioAvail} 
\end{table}

For the $12$ selected biomarkers, we estimated the disease timelines in the three aforementioned groups using the method selected after simulation experiments. We studied the positional variance of the estimated orderings by creating $100$ bootstrapped samples of the data.

\textbf{Experiment 4:} The above experiment was repeated by including volumes of subfields of hippocampus~\citep{Iglesias:2015} as biomarkers instead of considering the entire hippocampus volume as a single biomarker. We included the tail of the hippocampus (Tail), Subiculum, CA1, CA$2/3$,  hilar region of the dentate gyrus (CA4), hippocampal fissure (Fissure), area $27$ in~\cite{Brodmann:1909} (Presubiculum), area $49$ in~\cite{Brodmann:1909} (Parasubiculum), Molecular layer of the subiculum and CA fields (ML), granule cell and molecular layer of the dentate gyrus (GC\textunderscore ML\textunderscore DG), and the hippocampus-amygdala transition area (HATA). We excluded the white matter structures of Alveus and Fimbria in our analysis.

\textbf{Experiment 5:} In this experiment, we validated the disease stage $(\Upsilon_j)$ by computing its correlation with the subjects' MMSE and ADAS13 values. We used a $10$-fold cross validation, where the training set was used to estimate the disease timeline in the aforementioned groups and the test subjects' disease stage was evaluated by placing them on this disease timeline. We used the volume-based and CSF-based biomarkers from Experiment $3$, but excluded MMSE and ADAS13 scores from the model.

\begin{figure*}
\centering

        \begin{subfigure}[b]{0.33\textwidth}
        \centering
		\includegraphics[width=\textwidth]{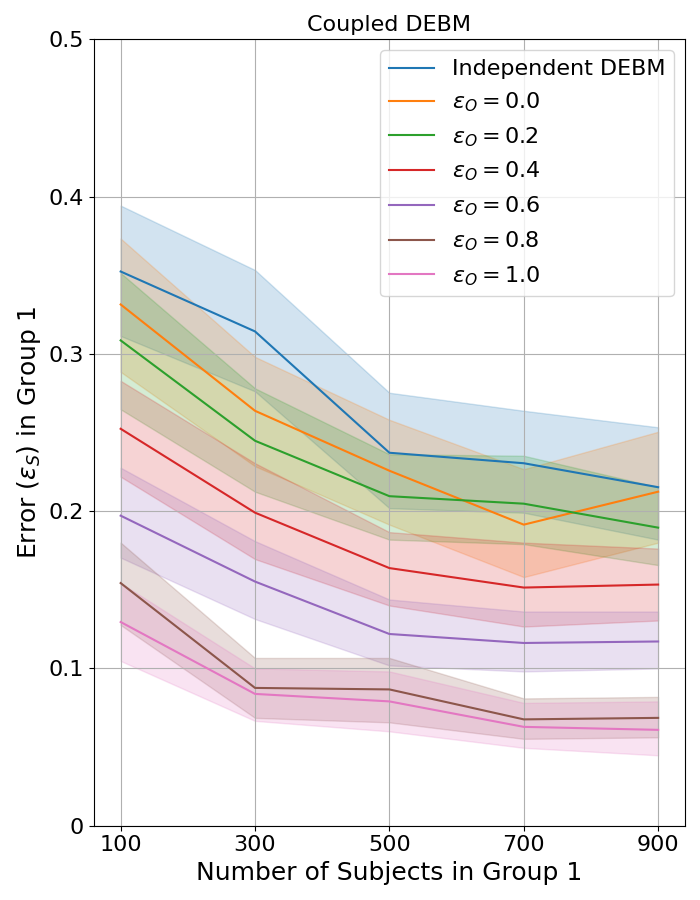}
		\caption[]{}
		\end{subfigure}
		\quad \quad 
		\begin{subfigure}[b]{0.33\textwidth}
		\centering
		\includegraphics[width=\textwidth]{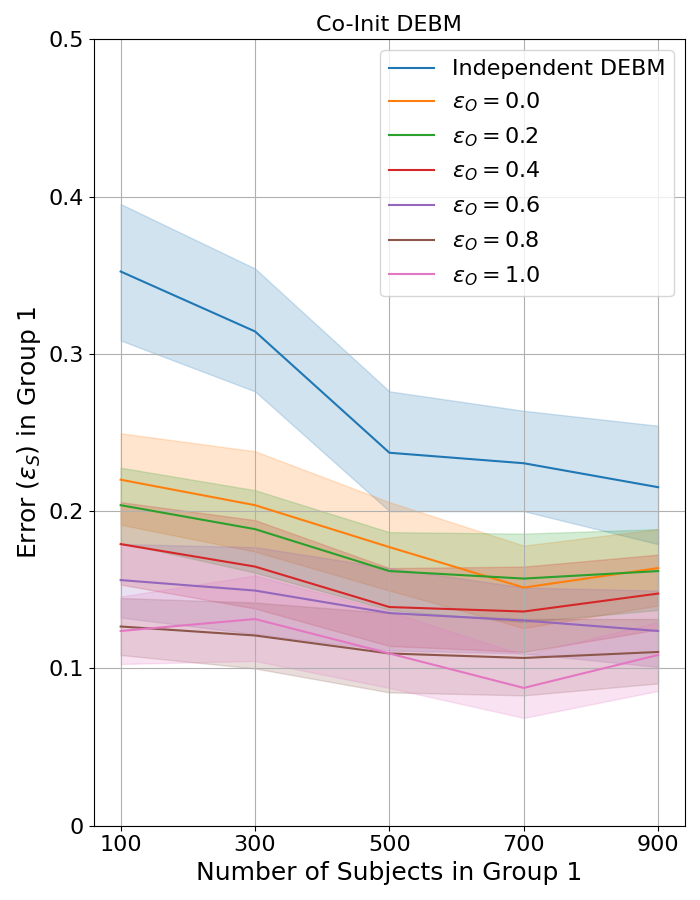}
		\caption[]{}
		\end{subfigure}
		
		\begin{subfigure}[b]{0.29\textwidth}
		\includegraphics[width=\textwidth]{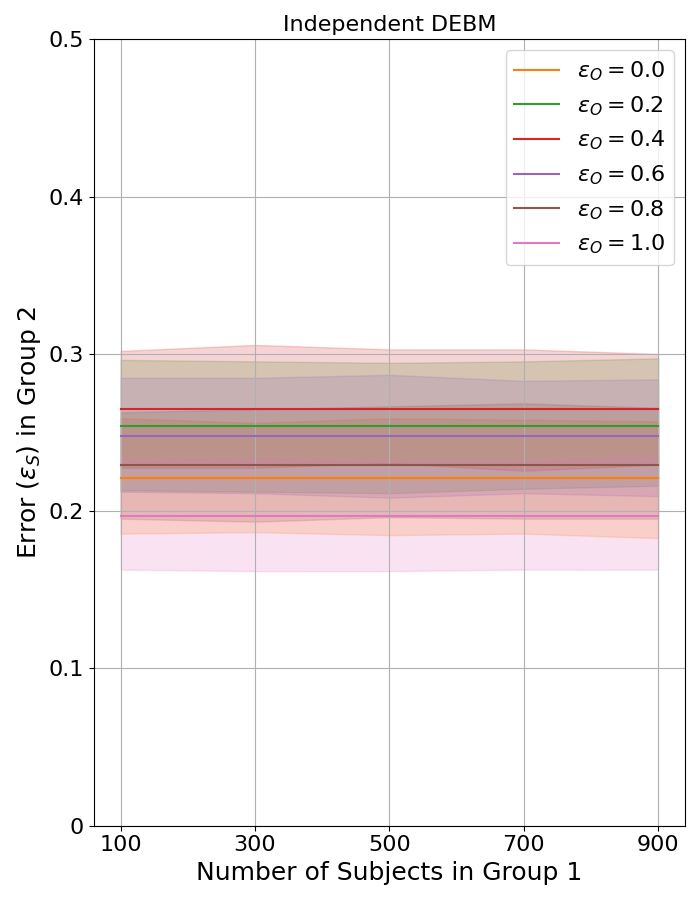}
		\caption{}
		\end{subfigure}
		\quad \quad 
		\begin{subfigure}[b]{0.29\textwidth}
		\includegraphics[width=\textwidth]{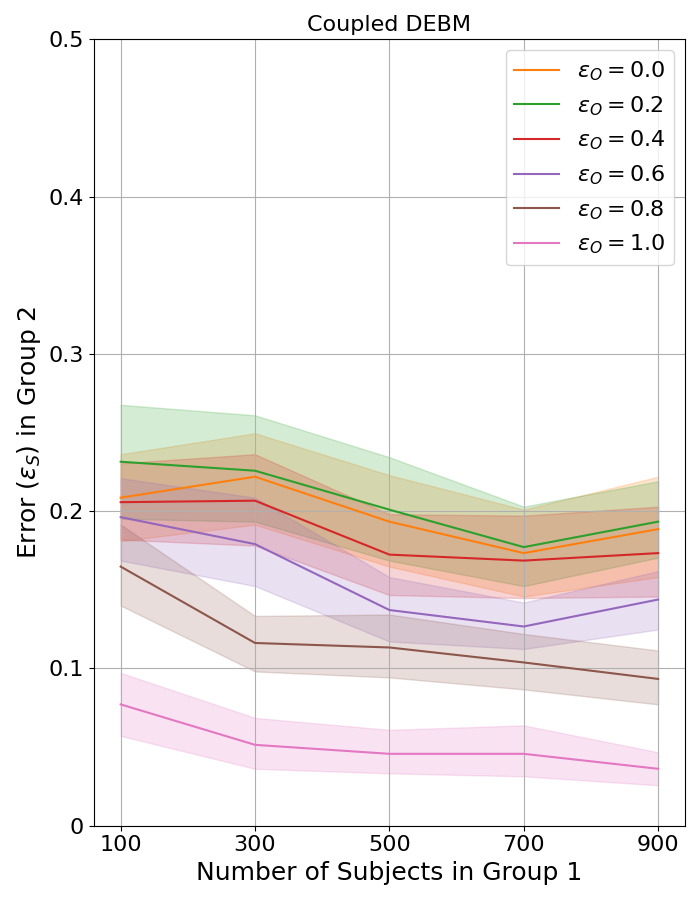}
		\caption{}
		\end{subfigure}
		\quad \quad 
		\begin{subfigure}[b]{0.29\textwidth}
		\includegraphics[width=\textwidth]{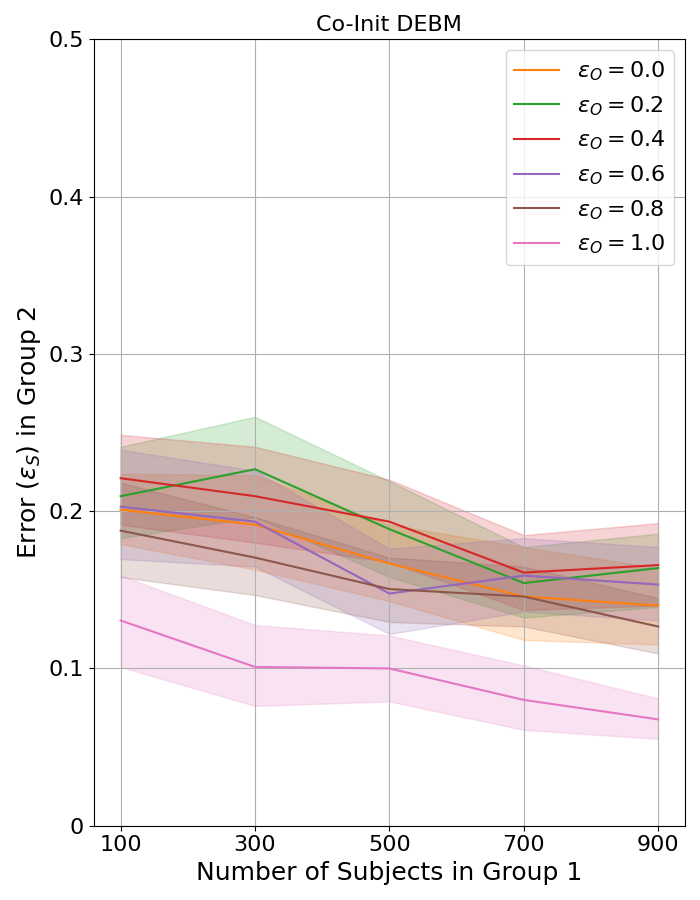}
		\caption{}
		\end{subfigure}
		
\caption{Experiment 1: The effect of $\varepsilon_O$ (the difference in groundtruth event orderings in the two groups) on the performance of the proposed methods. The shaded region in these plots represents standard deviation of the error in estimation of the proposed methods in $50$ random iterations of simulations. The plots in (a) and (b) show the ordering errors in Group $1$ using  Coupled DEBM and Co-init DEBM with independent DEBM shown in both (a) and (b), as a function of number of subjects in Group $1$. The plots in (c), (d) and (e) show the ordering errors in Group $2$ using independent DEBM, Coupled DEBM and Co-init DEBM respectively as a function of number of subjects in Group $1$.}
\label{fig:Exp1}
\end{figure*}

\begin{figure*}
\centering
        \begin{subfigure}[b]{0.3\textwidth}
            \centering
            \includegraphics[width=\textwidth]{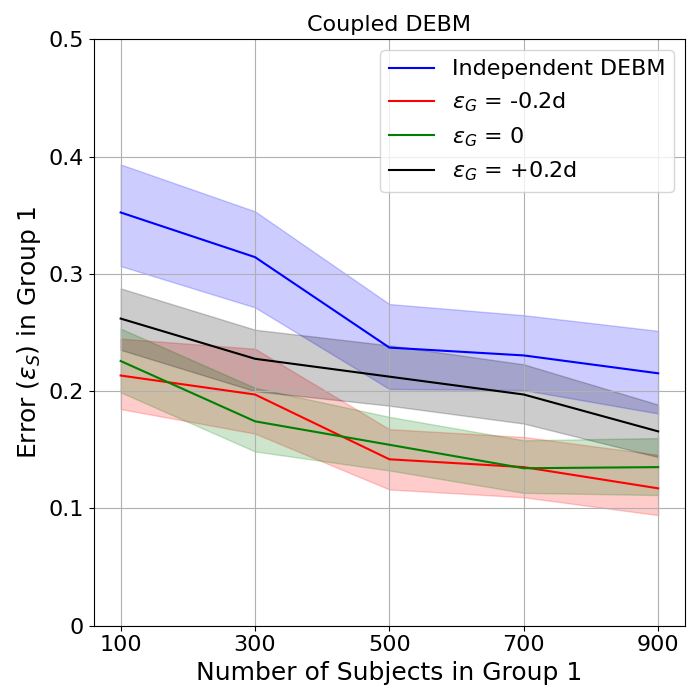}
            \caption{}    
        \end{subfigure}
        \quad \quad 
         \begin{subfigure}[b]{0.3\textwidth}
            \centering
            \includegraphics[width=\textwidth]{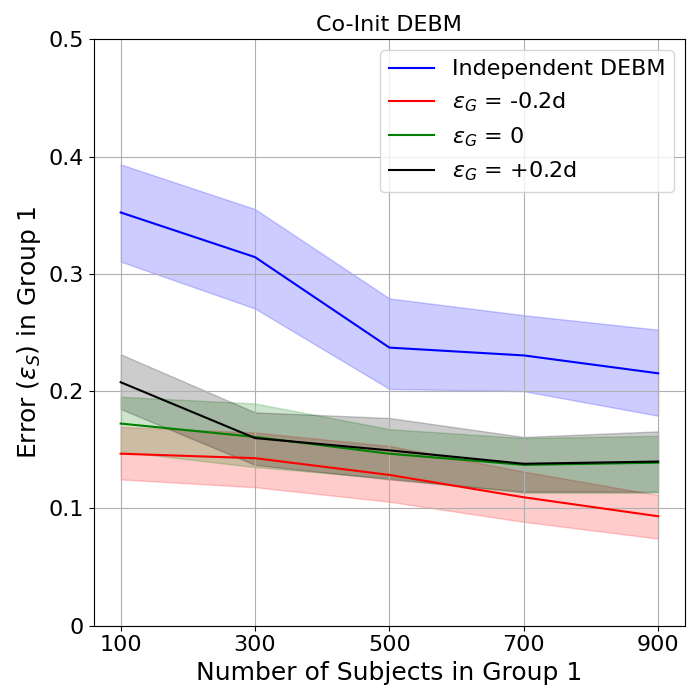}
            \caption{}    
        \end{subfigure}
        
        \begin{subfigure}[b]{0.3\textwidth}   
            \centering 
            \includegraphics[width=\textwidth]{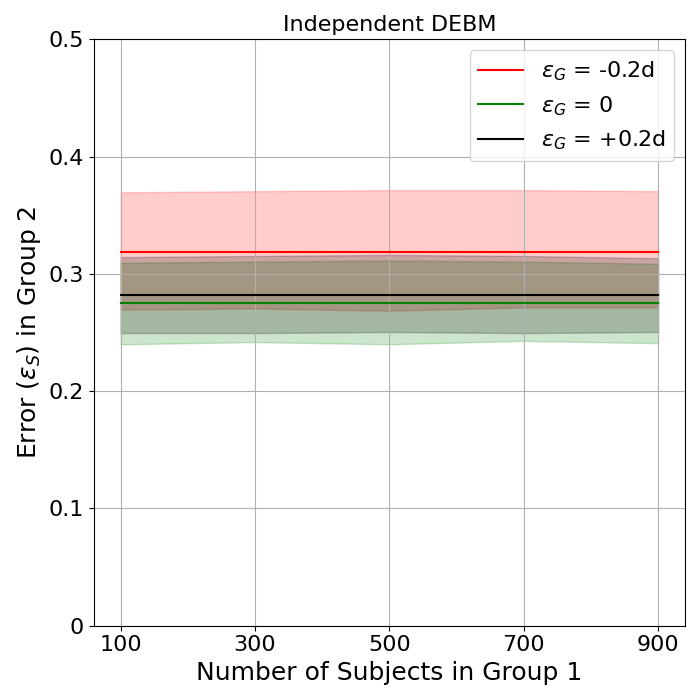}
            \caption {}    
        \end{subfigure}
        \quad \quad 
         \begin{subfigure}[b]{0.3\textwidth}   
            \centering 
            \includegraphics[width=\textwidth]{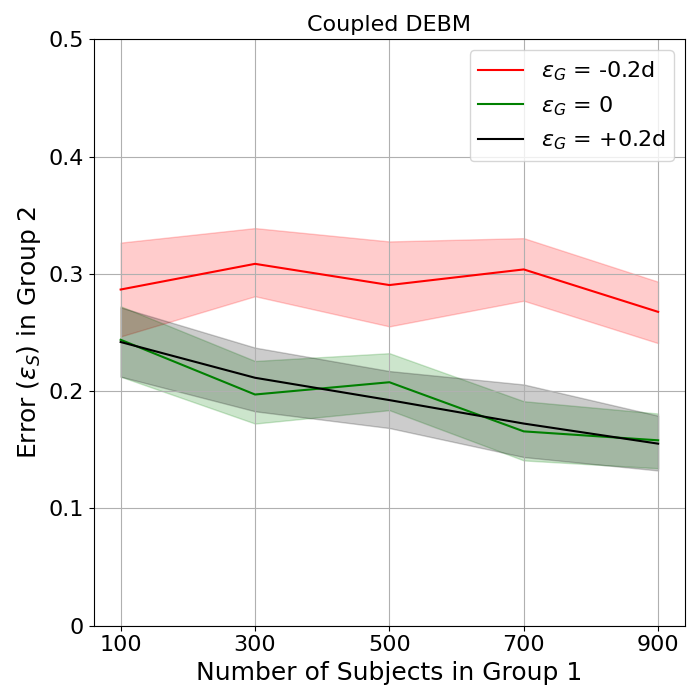}
            \caption {}    
        \end{subfigure}
        \quad \quad 
        \begin{subfigure}[b]{0.3\textwidth}   
            \centering 
            \includegraphics[width=\textwidth]{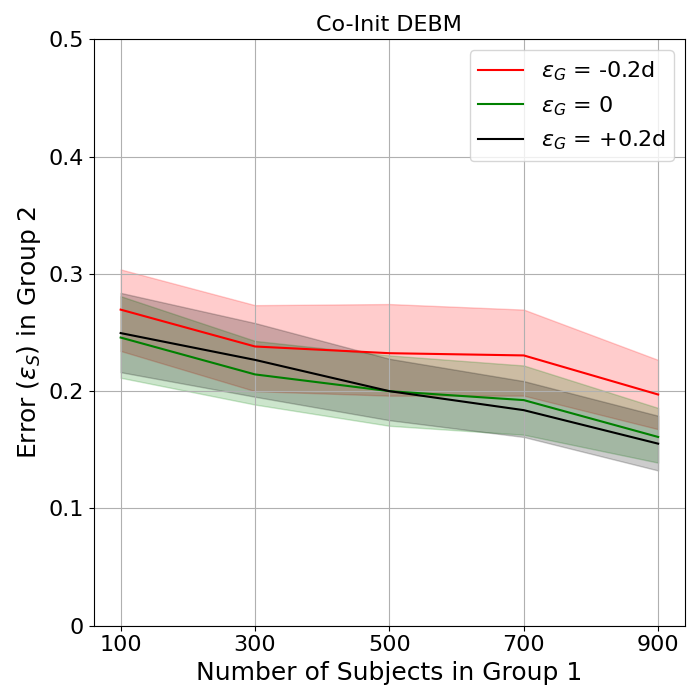}
            \caption {}    
        \end{subfigure}
\caption{Experiment 2: The effect of $\varepsilon_G$ (difference in abnormal biomarker levels in the two groups),    on the performance of the proposed methods. The shaded region represents standard deviation of the error in $50$ random iterations. The plots in (a) and (b) show the ordering errors in Group $1$ using Coupled DEBM and Co-init DEBM with independent DEBM shown in both (a) and (b), as a function of number of subjects in Group $1$. The plots in (c), (d) and (e) show the ordering errors in Group $2$ using independent DEBM, Coupled DEBM and Co-init DEBM respectively as a function of number of subjects in Group $1$.}

\label{fig:Exp2}
\end{figure*}

\begin{figure}[h!]
\centering
        \begin{subfigure}[b]{0.4\textwidth}
            \centering
            \includegraphics[width=\textwidth]{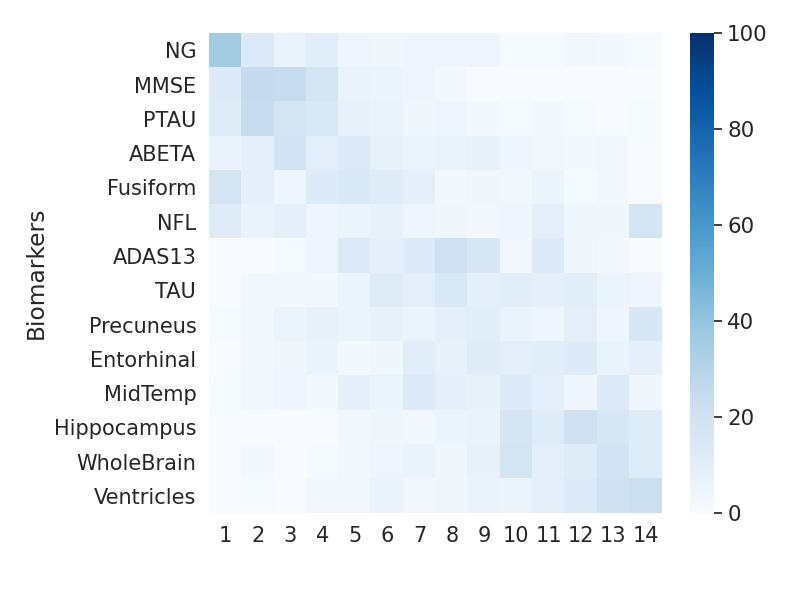}
            \caption[]%
            {{\small Protective group}}    
        \end{subfigure}

        \begin{subfigure}[b]{0.4\textwidth}   
            \centering 
            \includegraphics[width=\textwidth]{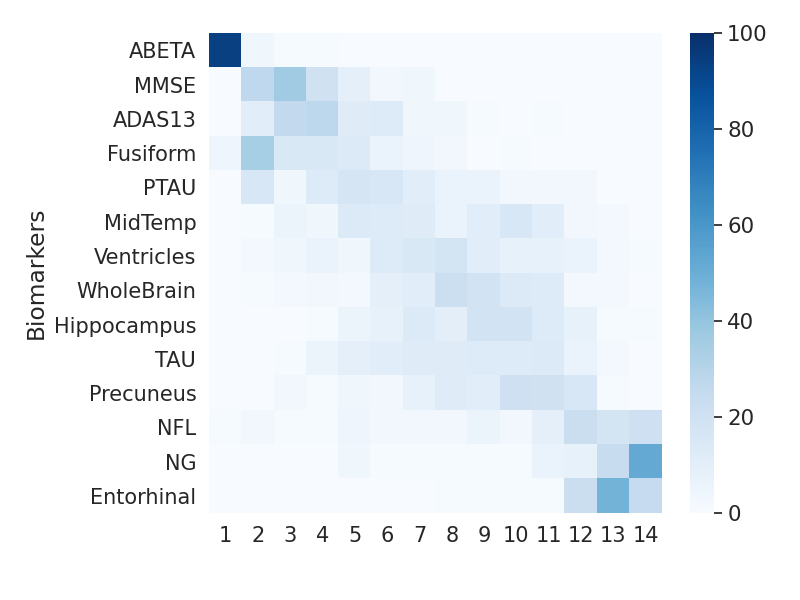}
            \caption[]%
            {{\small Neutral group}}    
        \end{subfigure}
 
        \begin{subfigure}[b]{0.4\textwidth}   
            \centering 
            \includegraphics[width=\textwidth]{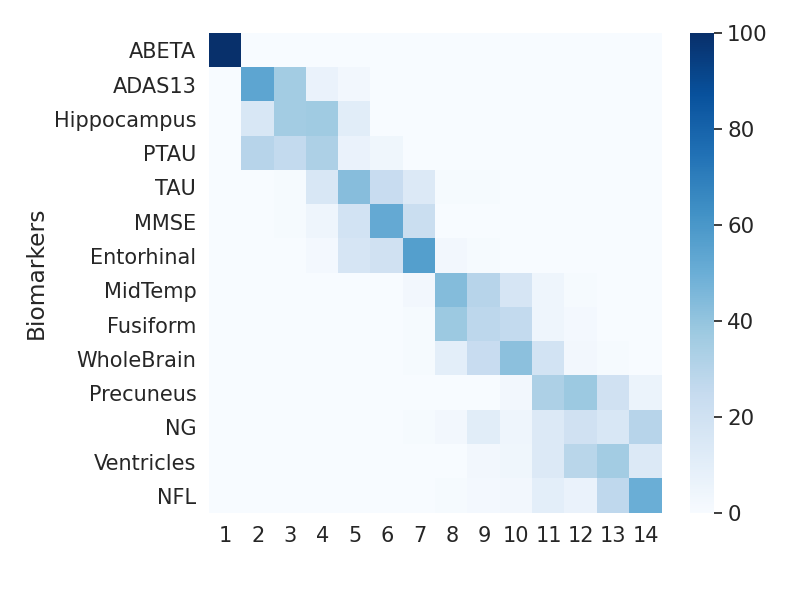}
            \caption[]%
            {{\small Risk group}}    
        \end{subfigure}
\caption{Experiment 3: Orderings of CSF, global cognition and volumetric biomarkers in the \textit{APOE} based protective, neutral and risk groups along with their uncertainty estimates. Uncertainty in the estimation of the ordering was measured by $100$ repetitions of bootstrapping, in the three \textit{APOE} based groups. The color-map is based on the number of times a biomarker is at a position in $100$ repetitions of bootstrapping. The number of subjects in the three groups were $75$, $411$ and $485$ respectively. The orderings were obtained using Co-init DEBM.}
\label{fig:Exp3}
\end{figure}

\begin{figure*}
\centering
        \begin{subfigure}[b]{0.45\textwidth}
            \centering
            \includegraphics[width=\textwidth]{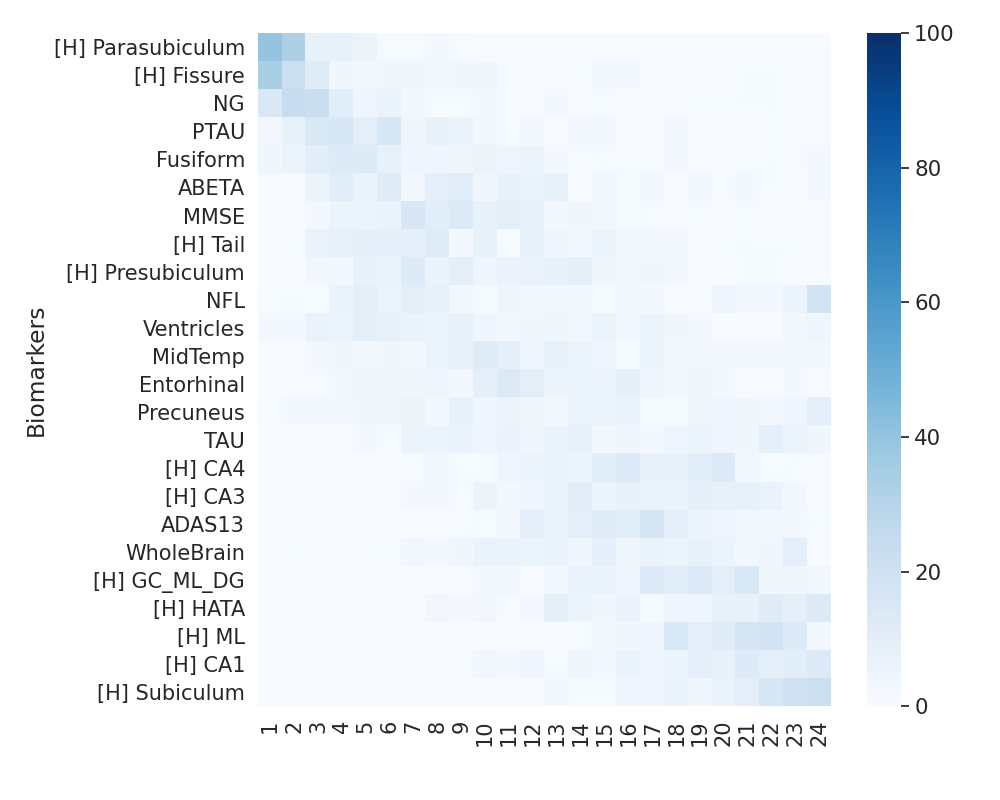}
            \caption[]%
            {{\small Protective group}}    
        \end{subfigure}

        \begin{subfigure}[b]{0.45\textwidth}   
            \centering 
            \includegraphics[width=\textwidth]{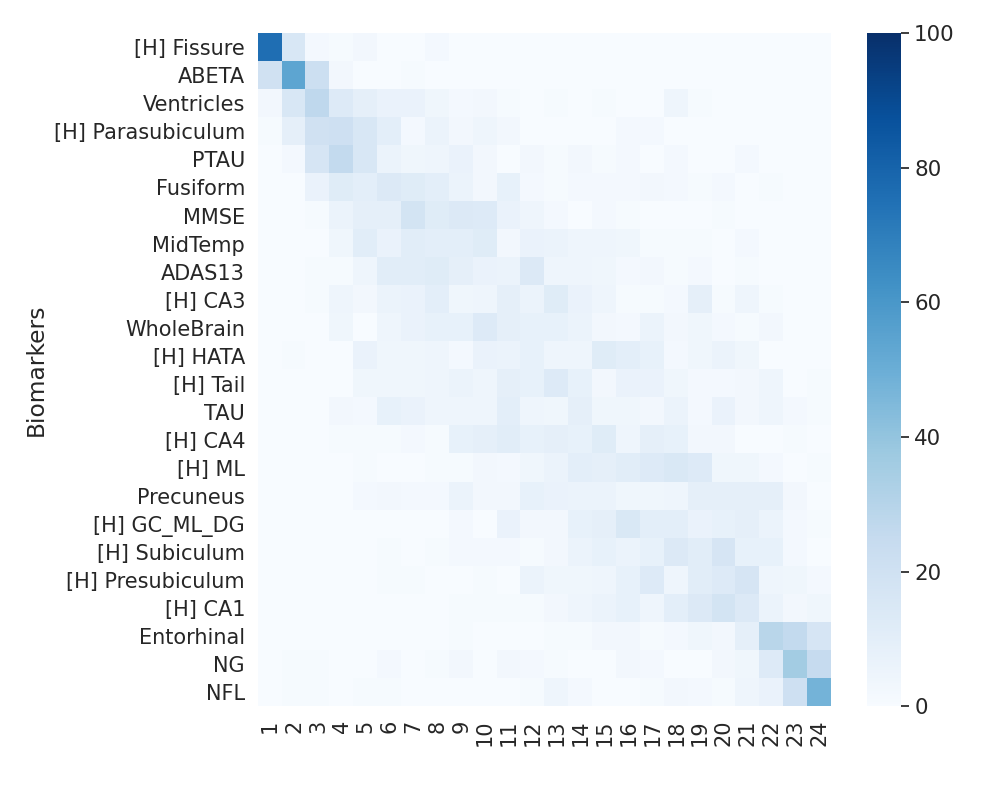}
            \caption[]%
            {{\small Neutral group}}    
        \end{subfigure}
 
        \begin{subfigure}[b]{0.45\textwidth}   
            \centering 
            \includegraphics[width=\textwidth]{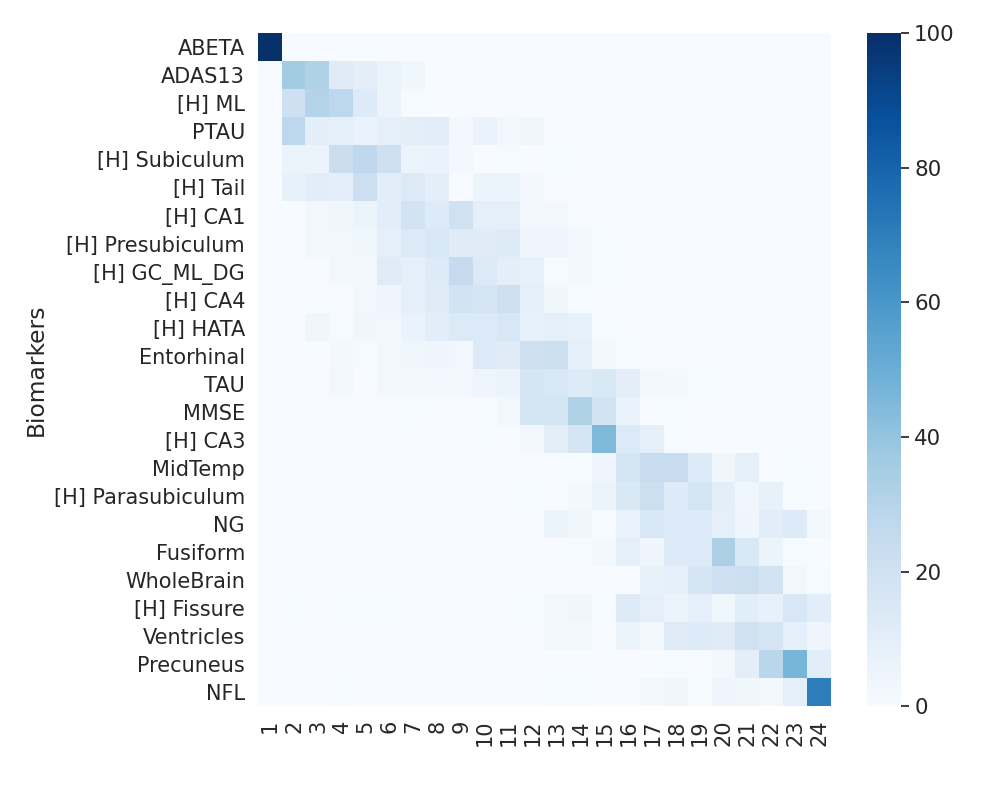}
            \caption[]%
            {{\small Risk group}}    
        \end{subfigure}

\caption{Experiment 4: Orderings of CSF, global cognition and volumetric biomarkers including those of Hippocampus subfields in the \textit{APOE} based protective, neutral and risk groups along with their uncertainty estimates. Uncertainty in the estimation of the ordering was measured by $100$ repetitions of bootstrapping, in the $3$ \textit{APOE} based groups. The color-map is based on the number of times a biomarker is at a position in $100$ repetitions of bootstrapping. The biomarkers for hippocampus subfields are indicated with a prefix of [H] in this figure. The number of subjects in the three groups were $75$, $411$ and $485$ respectively. The orderings were obtained using Co-Init DEBM.}
\label{fig:Exp4}
\end{figure*}

    \begin{figure*}
        \centering
        
        \begin{subfigure}[b]{0.325\textwidth}
            \centering
            \includegraphics[width=\textwidth]{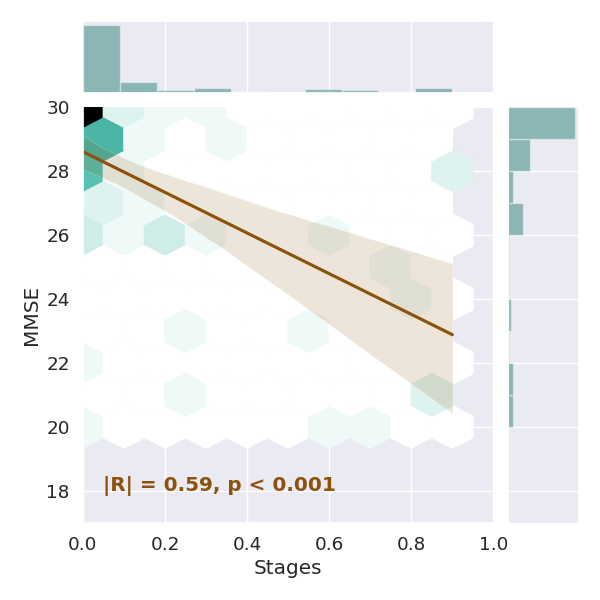}
            \caption[]%
            {{\small Protective: Stage vs MMSE}}    
        \end{subfigure}
        \quad \quad
        \begin{subfigure}[b]{0.325\textwidth}  
            \centering 
            \includegraphics[width=\textwidth]{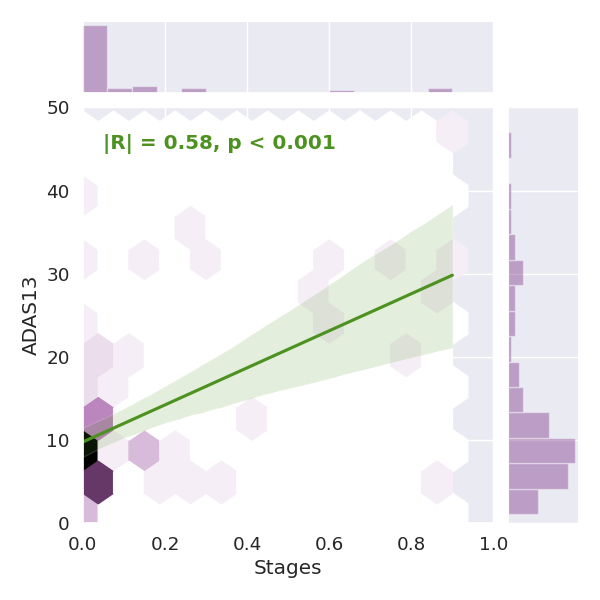}
            \caption[]%
            {{\small Protective: Stage vs ADAS13}}    
        \end{subfigure}
        
        \vskip\baselineskip
        \begin{subfigure}[b]{0.325\textwidth}   
            \centering 
            \includegraphics[width=\textwidth]{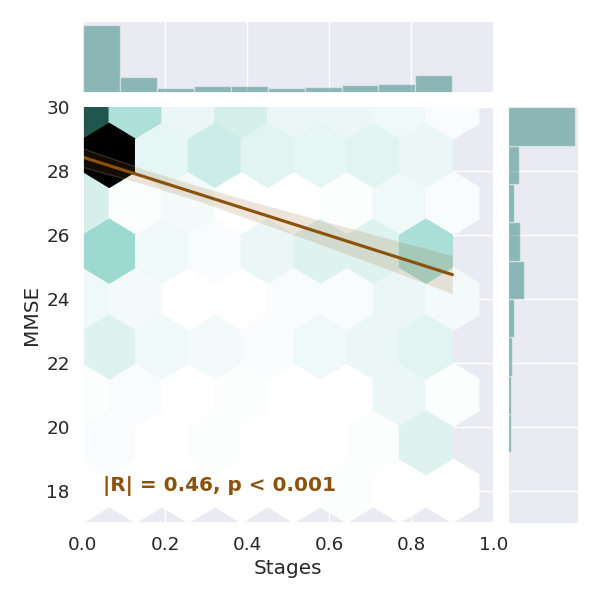}
            \caption[]%
            {{\small Neutral: Stage vs MMSE}}    
        \end{subfigure}
        \quad \quad
        \begin{subfigure}[b]{0.325\textwidth}   
            \centering 
            \includegraphics[width=\textwidth]{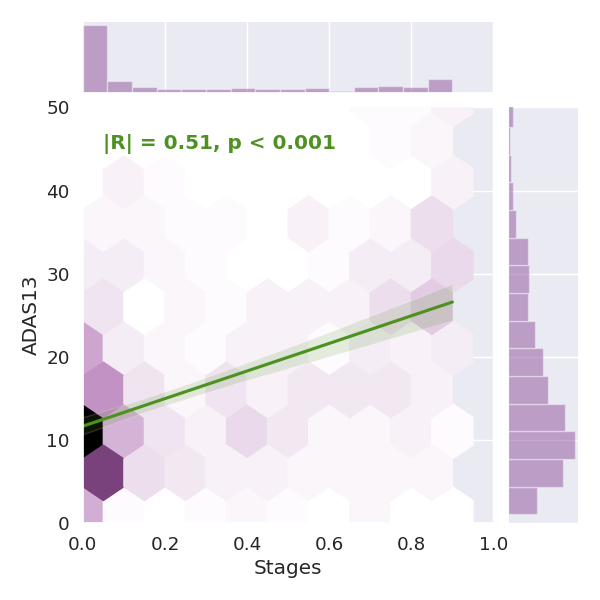}
            \caption[]%
            {{\small Neutral: Stage vs ADAS13}}    
        \end{subfigure}
        
        \vskip\baselineskip
        \begin{subfigure}[b]{0.325\textwidth}   
            \centering 
            \includegraphics[width=\textwidth]{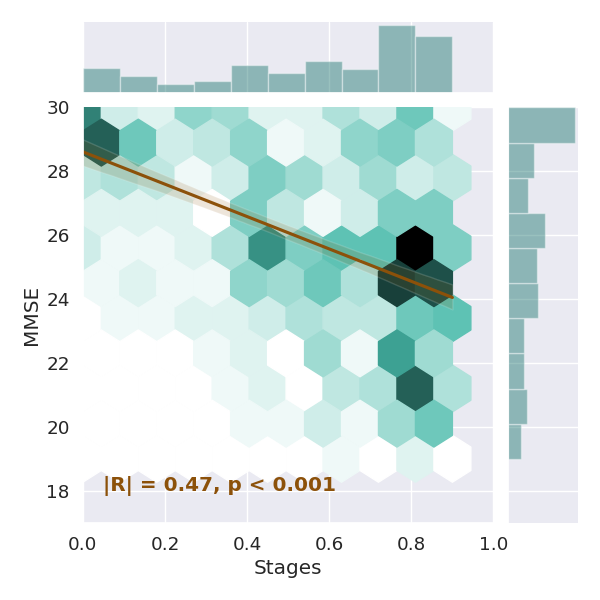}
            \caption[]%
            {{\small Risk: Stage vs MMSE}}    
        \end{subfigure}
        \quad \quad
        \begin{subfigure}[b]{0.325\textwidth}   
            \centering 
            \includegraphics[width=\textwidth]{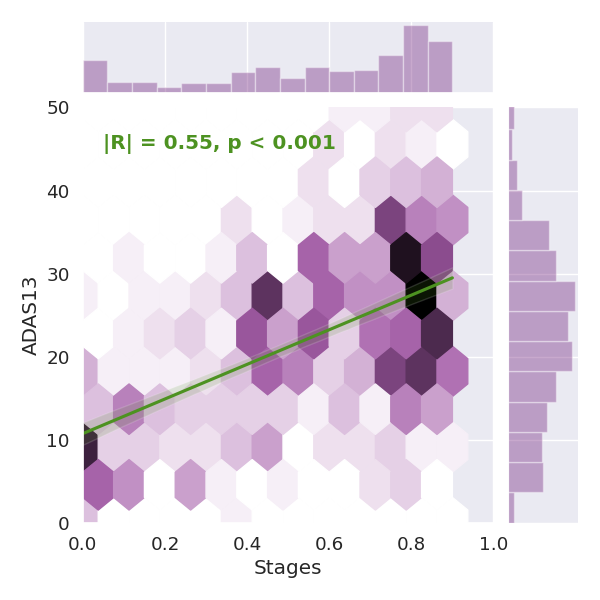}
            \caption[]%
            {{\small Risk: Stage vs ADAS13}}    
        \end{subfigure}
        
        \caption[]
        {\small Experiment 5: Correlation of estimated disease stages with MMSE and ADAS scores in the \textit{APOE} based protective, neutral and risk groups. The plot on top of each subfigure shows the probability density function of the disease stages, and the plot on the right of each subfigure shows the probability density function of the cognitive score in the subfigure. The 2D plot in each subfigure shows the joint density function of the two axes. The line in each subfigure shows the linear regression of MMSE / ADAS scores with the estimated disease stage and the shaded area around the line shows its $95\%$ confidence interval. Figures (a),(c) and (e) depict correlation between MMSE score and obtained disease stages in the three \textit{APOE} based groups. Figures (b), (d) and (f) depict correlation between ADAS13 score and the obtained disease stages in the three \textit{APOE} based groups.} 
        \label{fig:Exp5c}
    \end{figure*}
    
\section{Results} \label{sec:results}

\subsection{Simulations}

\textbf{Experiment 1:} Figures~\ref{fig:Exp1} (a) and (b) show the ordering errors $(\varepsilon_S)$ in Group 1 of the simulation datasets for DEBM, coupled DEBM and co-init DEBM as a function of number of subjects in Group 1, when $\varepsilon_O$ between the two groups changes from $0$ to $1$. Figures~\ref{fig:Exp1} (c), (d) and (e) show $\varepsilon_S$ in Group 2 of the simulation datasets for the aforementioned methods, as a function of number of subjects in Group 1. In our experiments, Group 1 dataset remains the same while Group 2 dataset changes as $\varepsilon_O$ increase. Hence DEBM results do not change with change in $\varepsilon_O$ in Figure~\ref{fig:Exp1} (a) and (b), whereas in Figure~\ref{fig:Exp1} (c), DEBM results do not change with increase in number of subjects in Group 1.

It can be seen that both coupled-training methods (i.e., co-init DEBM and coupled DEBM) outperform the default method of independently training DEBM models. It can also be observed that in both co-init DEBM and coupled DEBM the ordering errors decrease as $\varepsilon_O$ increases and that co-init DEBM outperforms coupled DEBM for lower values of $\varepsilon_O$, whereas the performance is on par with coupled DEBM for higher values of $\varepsilon_O$.

\textbf{Experiment 2:} Figures~\ref{fig:Exp2} (a) and (b) show $\varepsilon_S$ in Group 1 and Figures~\ref{fig:Exp2} (c), (d) and (e) show the same in Group 2, when varying $\varepsilon_G$. Even with $\varepsilon_G \neq 0$, coupled training (i.e., co-init DEBM and coupled DEBM) outperformed the default method of independently training DEBM models. Co-init DEBM showed negligible change in the errors when $\varepsilon_G \neq 0$. The performance of coupled DEBM in Group 1 worsened for $\varepsilon_G = +0.2d$ (Figure~\ref{fig:Exp2} (a)) and in Group 2 for $\varepsilon_G = -0.2d$  (Figure~\ref{fig:Exp2} (d)).

\subsection{Studying the effect of \textit{APOE}} 

The results in Experiments $1$ and $2$ show that the performance of co-init DEBM is more accurate and robust than coupled DEBM in most scenarios. We hence analyzed Experiments $3-5$ using co-init DEBM.

\textbf{Experiment 3:} Figure~\ref{fig:Exp3} shows orderings of CSF, global cognition and volumetric biomarkers in the \textit{APOE} based protective, neutral and risk groups along with their uncertainty estimates. It can be seen that the uncertainty of the ordering in the protective group was high. Despite this uncertainty, some biomarkers (i.e. MMSE, NG and PTAU) seem to occur earlier than the other biomarkers in this group. 

In the group of subjects with neutral alleles, ABETA was very prominently the earliest biomarker, followed by cognitive scores of MMSE and ADAS13. Among the CSF biomarkers, PTAU followed immediately after ABETA, which was inturn followed by TAU. NFL and NG were late biomarkers. Among the structural biomarkers, volumes of fusiform and middle-temporal gyri were the first to become abnormal, followed by ventricular volume and wholebrain volume. Hippocampus, precuneus and entorhinal volumes were late biomarkers in this group.

In the risk group, the CSF biomarkers followed a pattern that was similar to that of the neutral alleles. The cognitive biomarkers were early biomarkers in this group as well. However the ordering in structural biomarkers was very different from that in the neutral group. Hippocampus and entorhinal volumes were early biomarkers in this group, followed by middle-temporal and fusiform gyri volumes. Wholebrain, ventricular and precuneus volumes were late biomarkers.

\textbf{Experiment 4:} Figure~\ref{fig:Exp3} shows orderings of CSF, global cognition and volumetric biomarkers along with that of Hippocampus subfields in the \textit{APOE} based protective, neutral and risk groups along with their uncertainty estimates. Parasubiculum was an early biomarker in the protective group, which became abnormal even before MMSE and NG. Fissure, Tail and Presubiculum followed immediately after that. Subiculum was one of the late biomarkers.

Fissure and Parasubiculum were early structural biomarkers for the neutral group as well. This was followed by dysfunction in CA3 and HATA.

A majority of the Hippocampal subfields were early biomarkers in the risk group, with ML, Subiculum and Tail as the three earliest regions.

\textbf{Experiment 5:} The variation of MMSE and ADAS13 scores with respect to the estimated disease stages has been plotted in Figure~\ref{fig:Exp5c}, for all three groups. The patient stages showed a significant correlation with both MMSE and ADAS13 scores. The correlation coefficients were also comparable in the three groups.

\section{Discussion} \label{sec:disc}

DEBM models have been shown to be effective in determining the temporal cascade of biomarker abnormality as AD progresses, from cross-sectional data. In this work, we introduced a novel concept of splitting the different steps of DEBM into group-specific and group-aspecific parts for coupled training in stratified population. We considered two novel variations to split the steps of DEBM in this manner and through thorough experimentation in simulation datasets we observed that co-init DEBM helps in obtaining more accurate orderings in a  stratified population. Using this method, we estimated the biomarker cascades in AD progression with protective, neutral and risk alleles of \textit{APOE}, based on cross-sectional ADNI data. While the findings in neutral and risk groups fit the amyloid cascade hypothesis of Alzheimer's disease with high-confidence (which posits that ABETA accumulation triggers neurodegeneration), the finding in the protective group shows evidence for an alternative pathway (with relatively low confidence). In this section, we discuss the insights provided by the simulation experiments (Section~\ref{ssec:simdis}) used for method selection as well as the insights into the AD progression pathways provided by our experiments on the ADNI dataset (Section~\ref{ssec:apoedis}).


\subsection{Choice of the method}  \label{ssec:simdis}

Coupled DEBM and co-init DEBM both split DEBM into group-specific and group-aspecific steps for coupled training of an EBM in stratified populations. Experiment $1$ and $2$ showed that coupled training of the group-aspecific parts of DEBM and independently training the group-specific parts of DEBM results in more accurate orderings in the groups better than the default approach of independently training a DEBM model in each group.

While splitting DEBM into group-specific and group-aspecific parts, we started with the assumption that the latent true normal and abnormal biomarker distributions in the groups are either same or similar. The difference between co-init DEBM and coupled DEBM is that, co-init DEBM accounts for slight differences in the underlying biomarker distributions between the groups whereas coupled DEBM does not. 

The simulation dataset generated in Experiment $1$ had the same true normal and abnormal biomarker distributions in the different groups, from which the simulated subjects were randomly sampled, aligning well with the assumption of coupled DEBM. However, this did not result in overall better accuracies for coupled DEBM than that of co-init DEBM. Co-init DEBM was also more robust than coupled DEBM as its accuracy was less dependent on $\varepsilon_O$, the distance between the ground-truth orderings in the two groups.

Another observation in Experiment $1$, which was rather counter-intuitive, was that the errors made by the co-init and coupled DEBM models decreased as the distance between the ground-truth orderings in the two groups increased. When the orderings are further apart, the combined biomarker distributions in CN and AD groups have a larger overlap. The non-overlapping initialization (before the GMM optimization) thus results in the normal and abnormal distributions to be further apart. We hypothesize that this results in a better estimation of the mixing parameters during GMM optimization and in-turn resulted in more accurate orderings, as mixing-parameters are dependent on the biomarker's position in the ordering.

In Experiment $2$, we checked the performance of our approaches when the assumption (true normal and abnormal biomarker distributions being same across groups) is violated in the dataset. This experiment showed that the orderings obtained using co-init DEBM are more robust to differences between the abnormal Gaussians across groups than those obtained with coupled DEBM. With coupled DEBM, the error increased in the group with weaker biomarkers i.e., Group 1 in the case of $\varepsilon_G=+0.2d$ and Group 2 in the case of $\varepsilon_G=-0.2d$. This shows that coupled DEBM introduces a systematic bias in the estimation of ordering that is detrimental to the group with weaker biomarkers. Co-init DEBM also showed a similar bias, but to a much lesser extent.

We hence selected co-init DEBM as the preferred approach for splitting and performed our analysis on ADNI dataset using this approach. We expect that this idea of splitting DEBM into group-specific and group-aspecific parts can be easily extended to the EBM introduced by~\cite{Fonteijn:2012}.

\subsection{Cascade of biomarker changes in the \textit{APOE} based groups} \label{ssec:apoedis}

Dividing the total population into groups based on \textit{APOE} carriership enabled us to create more phenotypically homogeneous groups~\citep{Sandra:2019}, each with potentially specific disease progression timeline. In this section, we discuss our results in these \textit{APOE} carriership based groups.

Our finding suggests that among the CSF biomarkers in the \textit{APOE} risk and neutral groups, ABETA abnormality is the earliest biomarker event followed by PTAU. This fits the amyloid cascade hypothesis of AD, which posits that ABETA accumulation triggers neurodegeneration~\citep{Hardy:2017}. It also confirms the need for preventing the accumulation of ABETA in high-risk patients. NFL and NG are late biomarkers in the neutral and risk groups, which suggests that axonal~\citep{Ashton:2019} and synaptic~\citep{Thorsell:2010} degeneration do not occur until very late in the disease process in these groups. NG being abnormal after PTAU and TAU in the neutral and risk groups is also consistent with the previous findings that Tau mediates synaptic damage in AD~\citep{Jadhav:2015}.

In the protective group, we found that the abnormal NG and PTAU are the earliest CSF events, even before ABETA becomes abnormal. This could hint at the existence of an alternative pathway for the formation of tau tangles in the brain before ABETA accumulation, as suggested in~\cite{Weigand:2019}, but needs more extensive validation.

Among the volumetric biomarkers, Entorhinal cortex is one of the early biomarkers in the risk group which is supported by the findings in\mbox{~\cite{Huijbers:2014}}, but is one of the last biomarkers to become abnormal in the neutral allele. Ventricular volume is a late biomarker in the risk group but it becomes abnormal quite early in the neutral group as also observed by~\cite{Nestor:2008}. Hippocampus volume is the earliest biomarker in the risk group, but is a relatively late biomarker in the neutral and protective groups. This suggests that incidence of hippocampal sparing AD~\citep{Ferreira:2017} could correlate with \textit{APOE} carriership. 

Volumetric biomarkers of large regions are generally insensitive to small changes in volume of subparts. Thus to find the earliest regions that are affected in AD progression, we analyzed the ordering of biomarker events while also including hippocampal subfields. Our results suggests that in the risk group,  molecular layer of the subiculum and CA fields (ML) followed by subiculum and the tail of the hippocampus are the earliest volumetric biomarker events. Tail of the hippocampus is not a histologically distinct region but consists of CA1-4 and dentate gyrus~\citep{Iglesias:2015}. Each of these regions as well as subiculum were observed to be associated to AD in~\cite{Parker:2019}. Hippocampal fissure was one of the last regions to become abnormal. In contrast, in the neutral group, subiculum and CA1 were the last regions to become abnormal. Parasubiculum was one of the earliest region to become abnormal in this group. Parasubiculum as an early biomarker was also observed in~\citep{Zhao:2019} where they reported a significant difference in its volume between CN and MCI. In the protective group, Parasubiculum, Fissure, Tail and Presubiculum were among the early volumetric biomarkers. Our findings thus highlight that Hippocampus subfields are differentially affected due to \textit{APOE} carriership.

The findings related to these orderings of biomarker events were validated by correlating the patient stages derived from these orderings with MMSE and ADAS13 scores. Patient stages of subjects in all three groups, when used as test-subjects in a cross-validated manner, showed a significant correlation ($p<0.001$) with these scores. These correlations validate our findings and suggest that these genotype-specific disease progression timelines could be used for patient monitoring.

\section{Conclusion and Future work} \label{sec:conc}

We conclude that co-init DEBM provides the best accuracy and robustness when estimating orderings in stratified populations. Future work on co-init DEBM can focus on extending the approach for high-dimensional imaging biomarkers~\citep{Venkatraghavan:2019b}. This work also provides groundwork for extending the method towards hypothesis-free, data-driven stratification of phenotypes.

We gained new insights into the disease progression timeline of AD in the \textit{APOE} based risk, neutral and protective groups of \textit{APOE}. While we observed that the estimated disease progression timelines in the risk and neutral groups fit the amyloid cascade hypothesis with high confidence, the estimated time lines in the protective group may suggest an alternative pathway for the formation of tau tangles in the brain before amyloid $\beta$ accumulation, albeit with relatively low condence. We expect that these genotype-specific disease progression timelines will benefit patient monitoring in the future, and may help optimize selection of eligible subjects for clinical trials.

\section*{Acknowledgement}

This work is part of the EuroPOND initiative, which is funded by the European Union's Horizon 2020 research and innovation programme under grant agreement No. 666992. E.E. Bron acknowledges support from the Dutch Heart Foundation (PPP Allowance, 2018B011), Medical Delta Diagnostics 3.0: Dementia and Stroke, and the Netherlands CardioVascular Research Initiative (Heart-Brain Connection: CVON2012-06, CVON2018-28).

Data collection and sharing for this project was funded by the Alzheimer's Disease Neuroimaging Initiative (ADNI) (National Institutes of Health Grant U01 AG024904) and DOD ADNI (Department of Defense award number W81XWH-12-2-0012). ADNI is funded by the National Institute on Aging, the National Institute of Biomedical Imaging and Bioengineering, and through generous contributions from the following: AbbVie, Alzheimer’s Association; Alzheimer's Drug Discovery Foundation; Araclon Biotech; BioClinica, Inc.; Biogen; Bristol-Myers Squibb Company; CereSpir, Inc.; Cogstate; Eisai Inc.; Elan Pharmaceuticals, Inc.; Eli Lilly and Company; EuroImmun; F. Hoffmann-La Roche Ltd and its affiliated company Genentech, Inc.; Fujirebio; GE Healthcare; IXICO Ltd.; Janssen Alzheimer Immunotherapy Research \& Development, LLC.; Johnson \& Johnson Pharmaceutical Research \& Development LLC.; Lumosity; Lundbeck; Merck \& Co., Inc.; Meso Scale Diagnostics, LLC.; NeuroRx Research; Neurotrack Technologies; Novartis Pharmaceuticals Corporation; Pfizer Inc.; Piramal Imaging; Servier; Takeda Pharmaceutical Company; and Transition Therapeutics. The Canadian Institutes of Health Research is providing funds to support ADNI clinical sites in Canada. Private sector contributions are facilitated by the Foundation for the National Institutes of Health (www.fnih.org). The grantee organization is the Northern California Institute for Research and Education, and the study is coordinated by the Alzheimer's Therapeutic Research Institute at the University of Southern California. ADNI data are disseminated by the Laboratory for Neuro Imaging at the University of Southern California. 
    



\bibliographystyle{model2-names.bst}\biboptions{authoryear}






\section*{\refname}
\bibliography{EBM}





\end{document}